%% file: main.tex
\documentclass[manuscript, screen]{acmart}
\usepackage{multicol}
\usepackage{multirow}

\AtBeginDocument{%
  \providecommand\BibTeX{{%
    \normalfont B\kern-0.5em{\scshape i\kern-0.25em b}\kern-0.8em\TeX}}}

\setcopyright{acmcopyright}
\copyrightyear{2018}
\acmYear{2018}
\acmDOI{XXXXXXX.XXXXXXX}

\acmConference[ACM COMPASS '23]{Make sure to enter the correct
  conference title from your rights confirmation emai}{Aug 16--19,
  2023}{Cape Town, SA}
\acmPrice{15.00}
\acmISBN{978-1-4503-XXXX-X/18/06}

\begin{document}

\title{Hate Speech Detection in Limited Data Contexts using Synthetic Data Generation}

\author{Aman Khullar}
\authornote{Both authors contributed equally to this research.}
\email{amankhullar@gatech.edu}
\author{Daniel Nkemelu}
\authornotemark[1]
\email{dnkemelu@gatech.edu}
\affiliation{%
  \institution{Georgia Institute of Technology}
  \city{Atlanta}
  \state{Georgia}
  \country{USA}
}

\author{Cuong V. Nguyen}
\affiliation{%
  \institution{Georgia Institute of Technology}
  \city{Atlanta}
  \country{USA}}
\email{johnny.nguyen@gatech.edu}

\author{Michael L. Best}
\affiliation{%
  \institution{Georgia Institute of Technology}
  \city{Atlanta}
  \country{USA}
}

\renewcommand{\shortauthors}{Khullar and Nkemelu, et al.}

\newcommand{\compass}[1] {\textcolor{blue}{#1}}
\begin{abstract}
   A growing body of work has focused on text classification methods for detecting the increasing amount of hate speech posted online. This progress has been limited to only a select number of highly-resourced languages causing detection systems to either under-perform or not exist in limited data contexts. This is majorly caused by a lack of training data which is expensive to collect and curate in these settings. In this work, we propose a data augmentation approach that addresses the problem of lack of data for online hate speech detection in limited data contexts using synthetic data generation techniques. Given a handful of hate speech examples in a high-resource language such as English, we present three methods to synthesize new examples of hate speech data in a target language that retains the hate sentiment in the original examples but transfers the hate targets. We apply our approach to generate training data for hate speech classification tasks in Hindi and Vietnamese. Our findings show that a model trained on synthetic data performs comparably to, and in some cases outperforms, a model trained only on the samples available in the target domain. This method can be adopted to bootstrap hate speech detection models from scratch in limited data contexts. As the growth of social media within these contexts continues to outstrip response efforts, this work furthers our capacities for detection, understanding, and response to hate speech.
   \textbf{Disclaimer:} This work contains terms that are offensive and hateful. These, however, cannot be avoided due to the nature of the work.
\end{abstract}




\maketitle

\section{Introduction} \label{intro}
\input{Sections/intro}

\section{Related Work}
\input{Sections/relatedwork}

\section{Methodology}
\input{Sections/method.tex}

\section{Experiments \& Results} \label{exp}
\input{Sections/experiments.tex}

\section{Discussion}  \label{discussion}
\input{Sections/discussion.tex}

\section{Conclusion}
\input{Sections/conclusion.tex}

\begin{acks}
We thank Microsoft for providing compute for the experiment in Azure credits, and the Computing For Good Fellowship at Georgia Institute of Technology, which partially funded the first author’s work on this project. We would also like to thank our partners at The Carter Center for their support in the project. Finally, we would like to thank our Technologies and International Development Lab colleagues and the anonymous reviewers who provided critical feedback to help improve this paper.
\end{acks}

\bibliographystyle{ACM-Reference-Format}
\bibliography{main}


\end{document}

%% file: Sections/intro.tex
The increase in hateful content online has motivated research in automatic approaches for detecting hate speech~\cite{schmidt2019survey, fortuna2018survey}. Applied approaches from prior work have included heuristic (e.g., dictionaries, distance metrics, rule-based systems) and machine learning based (e.g., topic modeling~\cite{alshalan2020detection}, word embeddings~\cite{badri2022combining}, deep learning~\cite{ziqi2019hate}) methods. However, the task of detecting hate speech in limited data contexts is difficult~\cite{de2018hate,madukwe2022token}. There is a lack of datasets for training hate speech detection models in many languages and this presents one of the main long-standing challenges for hate speech detection~\cite{wu2020all}. This problem is exacerbated for less popular under-resourced languages~\cite{fortuna2018survey, jahan2023systematic}. 

Since only a small proportion of the huge amount of content generated daily is hate speech, most curated datasets have a very high class imbalance with a significantly small amount of positive hate class samples. Hate speech data collection and labeling tasks from scratch have shown to be expensive and not guaranteed to result in sufficient data for training a model~\cite{nkemelu2022tackling, whang2023data}. This work explores the effectiveness of synthetic data generation techniques for limited data contexts with little to no ground truth hate speech data. Within the scope of this paper, we describe high-resource languages as languages with ample availability of digital data broadly and hate speech data specifically. Limited data contexts refer to language domains with little to no labeled hate speech examples, whether or not they have unlabeled data resources. While these languages may be reasonably represented in language modeling data, they often do not have existing hate speech repositories to support the work of hate speech detection~\cite{joshi2020state}. These contexts represent the target context in this paper.

Data augmentation explores strategies for increasing the diversity of training samples without explicitly collecting new data \cite{feng2021survey}. Data augmentation techniques have increasingly been used for addressing imbalances or biases in training data by creating new data points through oversampling, heuristics, or geometric transformations~\cite{shorten2019survey}. This idea has been successfully applied in other domains, such as audio classification\cite{wei2020comparison}, and video classification \cite{yun2020videomix}. With considerations for the sensitive and subjective nature of hate speech, we draw on techniques from data augmentation to generate synthetic examples via context transfer from a freely available high-resource hate speech data repository to a language with limited hate speech data.

In this work, we address the issue of limited hate speech data by exploiting available resources from other higher resourced languages. We propose few-shot methods for hate speech data augmentation in limited data contexts. We then compared the performance of three synthetic hate speech generation methods. The first approach involves automatic machine translation (MT) of the hateful posts in a high-resource language to the limited data language. In the second approach, we identify suitable contextual replacement tokens in the hate speech examples from the high-resource language. Our method, contextual entity substitution (CES), takes as input a handful of examples in a language such as the English language, and heuristically replaces the person or group under attack in the high-resource context with potential hate-targeted persons/groups in the target context. This semi-heuristic method retains the sentiment of hate for the target group without altering the meaning of the text, as is prone in generative approaches. We then use an open-source language model, BLOOM \cite{scao2022bloom}, to synthetically generate hate speech examples in the target context. We design the prompts such that the model can generate hateful posts in the target context when given a few hate speech examples. 

We conducted multiple experiments to investigate the performance of the proposed data augmentation approaches in two languages: Hindi and Vietnamese. Though these languages are not considered low-resourced (since they are fairly represented in language modeling research due to their representation in unlabeled data sources such as Wikipedia), they have very little hate speech data available making it nonetheless difficult to train hate speech detection models~\cite{joshi2020state}. A systematic review of 463 hate speech research works found only 4\% and <1\% representation for the Hindi and Vietnamese languages~\cite{jahan2023systematic}. Our findings show that synthetic data generated via the contextual entity substitution (CES) method can further improve model performance on the target language. Our analyses indicate that the magnitude of the performance gain from CES is based on the careful curation of an entity replacement table that is sensitive to the quality of the replacement matching setup and domain drift.

In summary, the main contributions of this paper include the following:
\begin{enumerate}
    \item development of a method for employing synthetic data generation techniques to counter harmful content like hate speech on social media platforms especially in limited data contexts.
    \item empirical investigation of gains vs. noise trade-off in combining synthetic machine-translated hate speech data with few original hate speech posts from limited data contexts.
    \item development of a new use-case for multilingual large language models showing how generative language models can be used to develop models that counter hate speech.
\end{enumerate}

In the following sections, we present related work, explain our synthetic data generation methodologies in detail, present the experiments that we performed along with their results, and then discuss the implications of our results. The code, data, and the entity table used for our work is present in our GitHub repository \footnote{https://github.com/TID-Lab/synthetic\_dataset}.

%% file: Sections/relatedwork.tex
\subsection{Hate Speech Detection in Limited Data Contexts}
Detecting hate speech content in limited data contexts remains a critical yet challenging task for machine learning (ML) systems. Publicly-available ground truth datasets for hate speech, while abundant in some languages such as English and Chinese, are limited to nonexistent in other contexts such as Burmese and Tagalog. Data unavailability hampers the development of effective hate speech detection models in these contexts \cite{aluru2021deep, bigoulaeva2021cross}. Previous works have explored curating hate speech datasets in low-resource languages by leveraging the knowledge of context experts \cite{nkemelu2022tackling}. However, the data work required for curating hate speech datasets is often an expensive time-consuming step that is not guaranteed to return sufficient data for model training~\cite{nkemelu2022tackling, sambasivan2021everyone}. 

Earlier works have used SVMs, CNNs, and RNNs for hate speech and offensive language detection in limited data contexts~\cite{romim2021hate, chopra2020hindi, demilie2022detection}. With the growth of large language models, researchers have leveraged pre-trained multilingual language models such as BERT~\cite{devlin2018bert} and XLM-R~\cite{conneau2019unsupervised} to perform hate speech classification for limited data contexts via few-shot learning~\cite{ali2022hate, toraman2022large, aluru2021deep}. \citet{aluru2021deep} evaluated the effectiveness of the mBERT \cite{devlin2018bert} and LASER (Language-Agnostic SEntence Representations)~\cite{github_fb_laser} models in detecting hate speech content in both high-resource languages (such as English and Spanish) and low-resource languages (such as Indonesian and Polish) and found that the LASER embedding model with logistic regression performed best in the low-resource scenario, whereas BERT-based models performed better in the high-resource scenario. They also show that data from other languages tend to improve performance in low-resource settings.~\citet{Lauscher2020FromZT} also show that multilingual transformer models like mBERT tend to perform poorly in zero-shot transfer to distant target languages, and augmentation with few annotated samples from the distant language can help improve performance.

Other researchers have explored using transfer learning to adapt existing labeled hate speech data in English and other languages to unlabeled data in new target domains.  
This often involves leveraging cross-lingual contextual embeddings to make predictions in the low-resource language\cite{bigoulaeva2021cross, ranasinghe2021multilingual}. In their work, \citet{ranasinghe2021multilingual} analyzed how XLM-R, a cross-lingual contextual embedding architecture~\cite{conneau2019unsupervised}, performs on the task of detecting offensive language in languages such as Bengali and Hindi. They implemented a transfer learning strategy by sequentially training an XLM-R model on English-language offensive speech data, then on the offensive speech data of the lower-resourced language. They found that using the model fine-tuned on Hindi training data achieves an F1 score of 0.806, and fine-tuning on both Hindi and English training data yields an improved F1 score of 0.857.

Our work builds on these existing works by combining transfer learning techniques with contextual entity substitution and language generation methods. We employ a few-shot setup to train an mBERT model on some hate speech examples and then on the augmented data to measure improvement in model performance with synthetic data.

\subsection{Context Transfer Across Languages}
A more targeted approach to improve the performance of models on tasks in limited data contexts involves employing data from higher-resourced languages related to the limited data context. Exploiting similarity in vocabulary and syntax makes insights gained from the high-resource language data reasonably transferable to the limited data context~\cite{Khemchandani2021ExploitingLR, Xia2019GeneralizedDA}. \citet{Khemchandani2021ExploitingLR} proposed RelateLM, a mechanism to effectively incorporate new low-resource languages into existing pre-trained language models by aligning low-resource lexicon embeddings with their counterparts in a related high-resource language~\cite{Khemchandani2021ExploitingLR}. They tested the effectiveness of this mechanism on Oriya and Assamese, two Indic languages whose data are unavailable in the multilingual BERT model (mBERT). In contrast to monolingual BERT, they found benefits in starting from a BERT model fine-tuned on Hindi (a higher-resourced Indic language) and then using RelateLM to incorporate Oriya and Assamese. 

Within the context of hate speech, prior works have explored how models trained in one context can be transferred to a different language context~\cite{grondahl2018all, yoder2022hate}.~\citet{grondahl2018all} show that hate speech models tend to perform poorly on data that differ from their initial training data.~\citet{swamy2019studying} demonstrated that hate speech models trained on the BERT model tend to perform competitively for different datasets, though generalization depends highly on the training data used. In analyzing the generalizability of hate speech models,~\citet{yoder2022hate} found that targeted demographic categories such as gender/sexuality and race/ethnicity play a significant role and vary from one context to another. Our work takes a data-centric, rather than a model-centric approach. To address the generalization shortcomings of pretrained models, we focus on improving the synthetic data by transferring the hate sentiment to the limited data context and substituting the contextually relevant target of hate speech to create a new dataset that fits the new domain. 

\subsection{Data Augmentation and Synthetic Data Generation in NLP}
Recent advancements in the field of image generation \cite{goodfellow2020generative, ramesh2022hierarchical, saharia2022photorealistic, yu2022scaling}, text generation \cite{brown2020language, thoppilan2022lamda, raffel2020exploring} and speech synthesis \cite{oord2016wavenet, shen2018natural} have led to the development of an area of research in which model outputs can be used to retrain newer models. This helps reduce annotation costs, maintains data privacy, and can also help with data imbalance and scarcity issues. For audio processing, text-to-speech models are being used to provide the training data to reduce the word-error-rate of the speech recognition models \cite{hu2022synt++} and also help capture words that were not present in the training data \cite{fazel2021synthasr}. Image generation models are being used to improve the dermatology classifiers \cite{sagers2022improving}, detect floods \cite{cardoso2022conditional}, and action recognition \cite{kim2022transferable}. Techniques such as cropping and noise injection, are commonly applied in image and sound processing\cite{shorten2019survey, Perez2017TheEO}. However, these techniques do not work well for text data as they can potentially change the original meaning of the input sentence. To this end, there is a growing body of work on data augmentation for natural language processing exploring tasks such as machine translation \cite{Xia2019GeneralizedDA}, automatic speech recognition \cite{Meng2021MixSpeechDA}, and named-entity recognition \cite{Ding2020DAGADA}. 

Text generation models are helping mitigate the class imbalance problem by synthesizing new examples for classes with few-shot approaches \cite{lee2021neural}. 
A majority of these works frame the data augmentation requirement as a text generation task~\cite{Feng2021ASO}. For example, \citet{Xia2019GeneralizedDA} proposed a generalized framework for data augmentation for low-resourced machine translation by generating a parallel corpus between a given low-resourced language and English from a parallel corpus between a related high-resourced language and English through unsupervised machine translation \cite{Xia2019GeneralizedDA}. This technique increased model performance by 1.5 to 8.0 BLEU points compared to the supervised back-translation baseline.  The importance of diversity and naturalism has also been studied to help build better synthetic datasets \cite{baradad2021learning}.



Data augmentation techniques have successfully been applied to construct hate speech classifiers \cite{Hartvigsen2022ToxiGenAL, cao2020hategan}. For instance,~\citet{Hartvigsen2022ToxiGenAL} attempted to augment existing toxic content datasets by leveraging GPT-3, a text generation model, for generating large-scale data on toxic and benign statements targeted at minority identity groups. The authors found that not only was the machine-generated dataset of high quality, but toxicity detection models trained on it significantly outperformed those trained on existing human-curated toxicity datasets. Similar to these works, we aim to generate synthetic data to improve the hate-speech detection accuracy of the machine learning classifier. However, we situate our work specifically to improve hate speech detection accuracy for limited data contexts. We also present an alternative to large language models and provide a competitive synthetic data generation methodology through heuristic contextualization of hate speech for the low-resource language, which is taken from a high-resource language.

%% file: Sections/method.tex
In this section, we describe our methodology to augment hate speech posts in limited data contexts using synthetic data generation techniques and to evaluate their performance in model training. The initial step involves curating a hate speech dataset in a high-resource language, which is a relatively easy task and is described next.

\subsection{Dataset Curation} \label{data_curate}
To start the synthetic data generation process, the first step is to identify a high-resource language and curate hateful posts in the selected language. The sources for the hate speech dataset are diversified to mitigate bias or over-representation of a single target group or individual. This is a relatively easy task due to the abundance of such datasets in the high-resource context. For our experiments, we use English as our high-resource language and use data curated by \citet{mathew2021hatexplain}, which covers 18 different groups targeted with hate speech in the American context. The authors built a corpus of hate-speech posts using lexicons provided by \citet{davidson2017automated}, \citet{ousidhoum2019multilingual} and \citet{mathew2019spread}. To reduce ambiguity in the nature of the posts, we selected only posts labeled as \textit{hateful} and discarded posts labeled as \textit{offensive}. From this dataset, we use a subset of 3,000 hateful posts in English. We pre-processed this data to remove the tags, hashtags, links, and emoticons from the text. We consider only posts with a word length greater than two after this pre-processing step.


\subsection{Machine Translation} \label{amt}
After curating the hate speech posts, we use this data to augment the hate speech data in the target language. \citet{das2022data} show automatic machine translation can boost classification performance to detect hate speech in the limited data context. For our first synthetic data generation approach, we apply a similar methodology using Microsoft Azure's machine translation API to convert the curated hate speech posts into Hindi and Vietnamese. 


\subsection{Contextual Entity Substitution} \label{aug}
\begin{figure}
  \includegraphics[width=\linewidth]{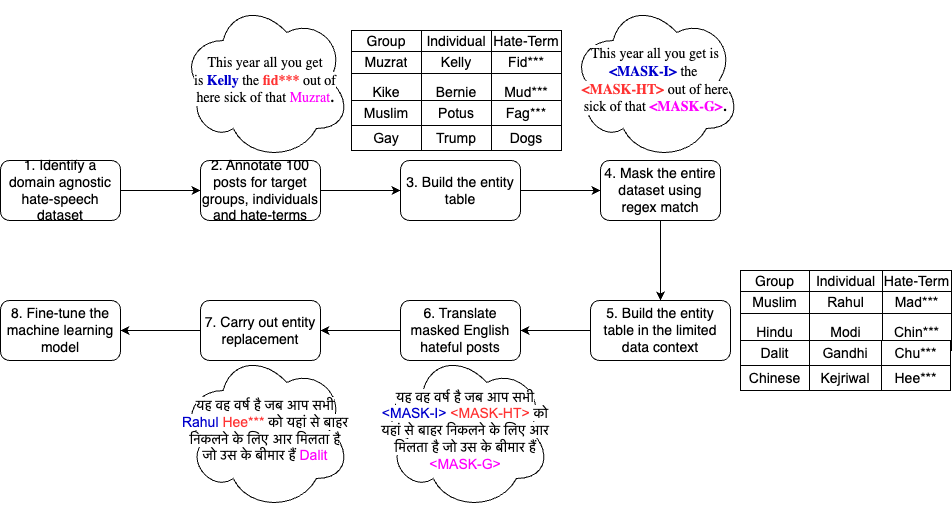}
  \caption{Framework to synthetically generate hate-speech posts in the limited data context. This framework takes the English hateful dataset as input and contextually translates it to the target language of interest with the help of human-curated entity tables.}
  \label{fig:synth_arch}
\end{figure}

Our second synthetic data generation approach builds on automatic machine translation. In this approach, we leverage the contextual nature of hate speech to account for differences in target groups and individuals based on different geography while transferring the hate sentiment across contexts. The main idea behind this approach is to identify the target entities subjected to hate speech or hate terms that are used in the high-resource context and substitute them with entities and hate terms from the target context. Figure \ref{fig:synth_arch} illustrates the framework we develop to generate synthetic hateful posts while accounting for this context shift.



The next step involves building an \textbf{\emph{entity table}} in the high-resource context. This entity table is an instantiation of the practice of creating lexicon lists as done in other works in the literature. For example, The PeaceTech Lab has curated hate
lexicons for languages spoken in conflict-prone countries such as Lebanon, Cameroon, and Sudan. The PeaceTech Lab Lexicons are a series of hate speech terms explaining inflammatory social media keywords and offering counter-speech suggestions to combat the spread of hate speech~\cite{peacetechlab}. However, these lexicons are only available for a handful of languages and contexts.

To create the entity table, we categorized lexicons into target groups, target individuals, hate terms, target countries, and political groups. We also differentiated entities (such as countries) that are present in the hateful posts but might not necessarily be the targets and created another category for them. We rely on multiple sources, including lexicons collected by \citet{mathew2020hate} which were derived from sources including Hatebase \footnote{https://hatebase.org} and the Urban Dictionary \footnote{https://www.urbandictionary.com}. We annotated 200 posts in the dataset to identify the most common target groups, individuals, countries, and hate terms and added them to the corresponding column in the entity table.



Subsequently, we automatically identify candidate entities for substitution in the hate speech dataset in the high-resource language. We adopt a heuristic approach that leverages the entity table and named entity recognition (NER) models. We iterate over the hate speech posts, find the words with a Levenshtein similarity score greater than a threshold value (0.75 in our case) to the words existing in the entity-table, and then replace these words with a corresponding \textit{MASK-x}. The \textit{MASK} corresponds to the entity we replace and the suffix \textit{x} represents the category of the entity. \textit{<MASK-G>, <MASK-I>, <MASK-CT>, <MASK-HT>, <MASK-P>} correspond to \textit{target groups, target individuals, target countries, hate-terms}, and \textit{political groups} respectively. For robust coverage in cases where certain names were not captured in our entity table, we used Spacy'S NER model \footnote{https://spacy.io/api/entityrecognizer} to identify all the entities with the tag \textit{PERSON} and replace it with \textit{<MASK-I>}

We created a similar entity table for the target context. To create this entity table, we ask two native speakers of Hindi and Vietnamese to review a sample of hate speech posts in their respective languages and to identify the hate target entities. Using this data and their experience with the context, they created the corresponding entity table for both languages. We subsequently included a lexicon of hate terms in the "hate-term" column of the entity table. This is the distinguishing part of the pipeline for different target contexts. We can create contextually relevant hateful posts in the target context of our interest just by modifying the contents of the entity table. Table \ref{tab:ent_table} shows the statistics of the entity table in English, Hindi, and Vietnamese.


After creating the entity table, we use the machine translation API to translate the masked English hateful posts into the target context. Our experiments showed that machine translation preserves the masks while translating the other words in the post. However, we also observed a slight loss in semantics during masked translation compared to standard translation. However, our study results show that the subsequent entity substitution was able to bridge this loss in semantic information, and the results are discussed further in section \ref{exp}.


\begin{table}[t]
  \centering
  \caption{Count of lexicons in the top three categories in the entity tables for English, Hindi, and Vietnamese}
  \label{tab:ent_table}
  \begin{tabular}{c|c|c|c}
    &English & Hindi & Vietnamese \\
    \hline
    Hate-terms& 56 & 19 & 23\\
    Target groups & 140 & 21 & 26\\
    Target individuals & 24 & 28 & 13\\
  \end{tabular}
\end{table}

Creating the synthetic hate speech posts involves combining the entity table and the masked translated posts in the target context. The different \textit{MASK-x} annotations help specify entity categories to replace to maintain semantic relevance. We randomly choose an entity from the corresponding entity category and replace the \textit{MASK-x} with that entity. We could theoretically increase the replacement seed to have an exponential number of synthetic hateful posts from a single masked translated post. However, through our initial experiments, we found that setting the seed value to 1 helps us get the best results---a reasonably diversified dataset that helps avoid overfitting.

The entity replacement step completes our pipeline for synthetically generating hateful posts in the limited data context. Table \ref{tab:qual_differences} shows how a machine-translated hateful post differs from a machine-translated hateful post with contextual entity substitution. In the first example in \ref{tab:qual_differences}, we find that the word ``kike'' has been transliterated from Latin to Devanagari. In contrast, we find that the CES method has the name of the person, ``Bhagat Singh'', referred to as ``penisless'', and is contextually relevant in the Indian domain (Indian freedom fighter). In the second example, we again find the word ``dyke'' being transliterated from the Latin script to the Devanagari script. The CES methodology, on the other hand, substitutes this entity with ``Heejra'', which stands for transgender people in India.

Finally, we use these synthetically generated posts to augment the hateful posts in the target context and then train the machine learning model for hate-speech detection. The results are explained in detail in section \ref{ces_exp}.

\begin{table}[h]
  \centering
  \caption{Qualitative difference between the synthetic hate-speech data using machine translation and the synthetic hate-speech data using contextual entity adaptation.}
  \label{tab:qual_differences}
  \begin{tabular}{c|p{0.3\linewidth}|p{0.3\linewidth}}
    S. No. & Machine Translated & Contextual Entity Substitution \\
    \hline
    1. & this ugly kike cunt keeps showing up on my timeline & this penisless Bhagat Singh keeps showing up on my timeline \\
    2. & angry bald dyke & angry bald Heejra \\
  \end{tabular}
\end{table}

\subsection{BLOOM Language Model}
Drawing from the advancements in natural language modeling, we investigate how large language models can help improve the classification accuracy of the hate-speech detection models. We develop a third methodology for generating synthetic hate-speech posts in the target context using the BLOOM language model with 3 billion parameters \cite{scao2022bloom}.

This method is different from the previous two methods as it depends only on a sample of hateful posts in the target context, which are used as few-shot examples to help the language model generate similar posts. To generate hate speech posts, we pass a few posts from the target context and prompt the language model to generate the sixth post as shown in table \ref{tab:bloom_lm}. The entire prompt is a string of Devanagari text given as input to the BLOOM-LM, and then it is asked to predict the following post from the input, which it predicts in Devanagari. Inferring from our empirical analyses, we chose the number of input examples as 5, set a repetition penalty of 2 to prevent post-repetition, applied early stopping with sampling, and specified a maximum token length of 100 words for the generated prompt. 



\begin{table}[t]
  \centering
  \caption{Prompt engineering to generate the hateful posts using the BLOOM Language Model}
  \label{tab:bloom_lm}
  \begin{tabular}{|p{\linewidth}|}
  \hline
    \begin{center} \textbf{Prompt} \end{center}
   \textbf{Post}: The fiscal deficit in previous governments was at an alarming level of 3.3 per cent. The situation was that no one was ready to give loan to UP because no one gives loan at a loss of more than 3 percent. Yogi ji reduced this deficit and brought it down to the level of 2.97 per cent.\\
    \textbf{Post}: Slogans of Pakistan indabad raised in Mumbai..!! Slogans of Pakistan Zindabad kept being raised in front of party's quota minister in Thackeray government, Abu Azmi! Respected Sir, it is a request that these Pakistan lovers, traitors, traitors should be badly thrown out of the country..\\
    \textbf{Post}: If this is the condition today, then tomorrow it will definitely be seen in UP and Delhi! Rather, people from every corner of the country are settled in Delhi, from where will they show their papers!\\
    \textbf{Post}: People call Yogi government as casteist, it is very shameful that they have always run governments for the health of one caste.\\
    \textbf{Post}: In the case of rape and subsequent brutal murder of Dr. Priyanka Reddy in Hyderabad, India's so-called secularists are refraining from raising their voice today because the accused Muslim and the locality, Asaduddin Owaisi, is it not enough to protest?\\
    \textbf{Post}:\\
    \hline
    \begin{center}\textbf{Generated Post}\end{center}
    \textcolor{black}{Do you know that India was going to become a world leader, but by ruining it by people like Modiji, we had become the poorest nation in the world.}\\
    \hline
  \end{tabular}
\end{table}

\subsection{Model and Metrics}
After generating the different types of synthetic data, we fine-tune the Multilingual BERT model \cite{devlin2018bert} using them. We report the average F1 scores of three independent runs of the training step. The same methodology is adopted for Hindi and Vietnamese.

%% file: Sections/experiments.tex


We focus on generating synthetic hateful posts to reduce the data imbalance problem and bootstrap hate speech detection work in new contexts. We collected datasets from high-resource and limited data contexts to perform our experiments. The dataset collected from the high-resource domain (i.e. English) supports the translation and entity substitution steps. The other datasets (in Hindi and Vietnamese) contain non-hateful posts and a small set of hateful posts on which data augmentation is performed.


\subsection{Training Data}
For Hindi, we use the dataset curated by \citet{bhardwaj2020hostility}, and for Vietnamese, we use the dataset curated by \citet{luu2021large}. Table \ref{tab:dist_tab} shows the distribution of the hateful and non-hateful posts in each of the datasets. Since we only use hateful posts from English, we report only the amount of hateful posts available in English. As illustrated in table \ref{tab:dist_tab}, the number of hateful posts was the least in the Hindi dataset. Hence, we keep 450 posts as the upper limit for our data in the low-resource language. Using a few-shot training setup, we gradually augment the hateful posts in the low-resource language with synthetic hateful posts.

\begin{table}[t]
  \centering
  \caption{Distribution of non-hateful and hateful posts in different data sources}
  \label{tab:dist_tab}
  \begin{tabular}{c|c|c|c|c|c|c}
    &  \multicolumn{2}{|c}{Train Set} & \multicolumn{2}{|c}{In-Domain Test Set} & \multicolumn{2}{|c}{Out-Of-Domain Test Set}\\
    \hline
    & Non-Hateful & Hateful & Non-Hateful & Hateful & Non-Hateful & Hateful \\
    \hline
    English & - & 5936 & - & - & - & - \\
    Hindi & 3050 & 478 & 277 & 133 & 1753 & 1017 \\
    Vietnamese & 19886 & 2556 & 4344 & 642 & - & - \\
  \end{tabular}
\end{table}


\subsubsection{Test Data}
To make our experimental conditions mirror real-world scenarios, our test dataset contains only original posts curated from the limited data context. We use the test data provided by \citet{bhardwaj2020hostility} and \citet{luu2021large} in Hindi and Vietnamese respectively. Since this test data is obtained from the same source as the training data, we call this an in-domain test set. However, in field deployments, we find real-time production data varies from the dataset on which the classifier was trained. This difference could be due to the different forms of hate speech on different social media platforms, domain and narrative shifts, or dissimilarity in data curation methodologies. To observe the performance of the trained models in such a scenario, we leverage another dataset in Hindi by \citet{bohra2018dataset} and term this the Out-Of-Domain (OOD) test set. This data comprises Hindi-English code-mixed posts in contrast to the training data, which comprised unilingual hate speech posts in Hindi. We transliterate this code-mixed data into Devanagari to carry out our test experiments. Due to the limited availability of open-source hate-speech datasets in Vietnamese, we did not perform the OOD analysis in Vietnamese.

\begin{table}[b]
  \centering
  \caption{Comparison of contextual entity substitution and BLOOM language model synthetic data generation methodologies vs. machine translated synthetic data generation methodology for Hindi (H) and Vietnamese (V). For each run, we use a constant 450 non-hateful posts.}
  \label{tab:synth_vs_trans_aug}
  \begin{tabular}{c|c|c|c|c|c}
    S. No. & Model Type & Original hateful posts & Synthetic hateful posts & Mean macro F1 (H) & Mean macro F1 (V) \\
    \hline
    1. & Base & 100 & 0 & 84.46 & 64.29 \\
    \hline
    2a. & All-Orig & 150 & 0 & 85.76 & 66.58 \\
    2b. & MT & 100 & 50 & 84.69 & \textbf{63.66} \\
    2c. & CES & 100 & 50 & \textbf{85.62} & 63.25 \\
    2d. & BLOOM-LM & 100 & 50 & 85.35 & 63.47 \\
    \hline
    3a. & All-Orig & 200 & 0 & 86.85 & 66.10 \\
    3b. & MT & 100 & 100 & 84.07 & 63.33 \\
    3c. & CES & 100 & 100 & 84.52 & 63.04 \\
    3d. & BLOOM-LM & 100 & 100 & \textbf{84.91} & \textbf{64.48} \\
    \hline
    4a. & All-Orig & 250 & 0 & 86.77 & 66.89 \\
    4b. & MT & 100 & 150 & 84.32 & \textbf{63.82} \\
    4c. & CES & 100 & 150 & \textbf{85.54} & 61.94 \\
    4d. & BLOOM-LM & 100 & 150 & 85.13 & 63.70 \\
    \hline
    5a. & All-Orig & 300 & 0 & 87.71 & 67.44 \\
    5b. & MT & 100 & 200 & \textbf{85.80} & 62.48 \\
    5c. & CES & 100 & 200 & 85.06 & 62.26 \\
    5d. & BLOOM-LM & 100 & 200 & 85.25 & \textbf{63.74} \\
    \hline
    6a. & All-Orig & 350 & 0 & 86.93 & 65.03 \\
    6b. & MT & 100 & 250 & 85.62 & 62.16 \\
    6c. & CES & 100 & 250 & \textbf{85.91} & 61.78 \\
    6d. & BLOOM-LM & 100 & 250 & 84.25 & \textbf{63.79} \\
    \hline
    7a. & All-Orig & 400 & 0 & 88.48 & 65.22 \\
    7b. & MT & 100 & 300 & 84.05 & 63.34 \\
    7c. & CES & 100 & 300 & \textbf{85.84} & 61.06 \\
    7d. & BLOOM-LM & 100 & 300 & 84.92 & \textbf{63.92} \\
    \hline
    8a. & All-Orig & 450 & 0 & 87.61 & 63.31 \\
    8b. & MT & 100 & 350 & 84.65 & 62.00 \\
    8c. & CES & 100 & 350 & \textbf{85.99} & 63.25 \\
    8d. & BLOOM-LM & 100 & 350 & 84.68 & \textbf{64.23} \\
  \end{tabular}
\end{table}

\subsection{Model Details}
For all experiments, we fine-tuned the cased multilingual BERT model~\cite{devlin2018bert}. We used BERT's sub-word tokenizer to tokenize the pre-processed input post and encode it into 768 dimensions using BERT embeddings. The encoding layer is followed by a dropout layer with a probability of 0.1, followed by a linear output layer which projects the 768-dimensional embedding into a 2-dimensional vector. We use the Cross-Entropy loss function and Adam optimizer to train the model. We use a batch size of 16, learning rate of $1e^{-05}$ without weight decay, gradient clipping norm of $1.0$, and fine-tuned the model for 10 epochs. We separate 10\% of the training dataset for cross-validation and use 90\% of the training data during the fine-tuning step. Our model has 177M trainable parameters, and we use a Microsoft Azure Virtual Machine with 1 GPU and 8GB memory to fine-tune the model. Below, we report the experimental setup and our results.

\subsection{Synthetic hateful augmentation through machine translation} \label{mt_exp}
We analyzed the impact of machine-translated hateful posts from English for training hate speech detection models in Hindi and Vietnamese. We use non-hateful posts available in Hindi and Vietnamese and a baseline of 100 original hateful posts in both languages. This mimics the typical real-world case where a handful of labeled hateful data is available compared to a majority of non-hateful posts. This initial split had 18\% of the training data with true labels as original hate speech and about 82\% for non-hate speech. This base case demonstrates a mean F1 score of 84.46 and 64.29 for Hindi and Vietnamese respectively. 

To test the effectiveness of machine-translated (MT) examples for augmentation, we increase the baseline training data by adding 50 original hateful posts to the training data and comparing the results with a training data setup of the baseline of 100 original hateful + 50 new machine-translated hateful posts. We iteratively execute this increment of original vs. synthetic for seven steps until the hateful/non-hateful split is even (50:50\%). Table \ref{tab:synth_vs_trans_aug} shows the Macro F1 score of models trained on the baseline and subsequent synthetic increments. The all-original model (All-Orig) acts as an upper limit to the performance if we had a complete set of original hateful posts and did not need to perform data augmentation. Our results show that in the ideal case where additional original hateful posts are added to the training data, the model performance attained F1 scores up to 88.48 (All-Orig, 7a) and 67.44 (All-Orig, 5a), compared to the initial baseline scores of 84.46 and 64.29 for Hindi and Vietnamese respectively.

\emph{Finding}. We observe that models trained using data augmented with machine-translated posts showed very little improvement on the baseline for Hindi (with mean F1 scores ranging from 84.05 - 85.80 vs. 84.46 baseline) but did not outperform the baseline for Vietnamese (with mean F1 scores ranging from 62.16 - 63.66 vs. 64.29 baseline). In general, the MT models did not significantly improve on the baseline as more translated data were added indicating that the MT data potentially introduced more noise and less signal to the model.

\subsection{Synthetic hateful augmentation through contextual entity substitution} \label{ces_exp}
We follow the setup described in \ref{mt_exp} to compare the baseline results with synthetic examples generated from the original English hate speech dataset using our contextual entity substitution method (CES) described in \ref{aug}. Similarly, we use 450 non-hateful posts and iteratively augment the original hateful posts in Hindi and Vietnamese with synthetic CES posts in increments of 50. Table \ref{tab:synth_vs_trans_aug} shows the comparative results between the contextual entity substitution method (CES) vs. the MT and All-Orig models for Hindi and Vietnamese. 

\emph{Finding}. For Hindi, we find that in the majority of the steps, the CES methodology outperforms the machine-translated methodology and closes the gap with the models trained on all original hateful posts of the same quantity. The CES methodology shows a boost in performance with a mean F1 score up to 85.99 with 350 synthetic hate posts (CES, 8c) which is better than both the performance of the baseline of 100 original hate posts alone and MT-augmentation for all cases. For Vietnamese, both the MT and CES scenarios show a decrease in performance after adding more synthetic data resulting in mean F1 scores that were lower than the baseline. Broadly, we observe an increase in performance for CES-augmented models in Hindi. However, there is a surprising dominance of MT over CES methods in Vietnamese. We hypothesize that this is possibly due to the nature of the entity table for Vietnamese and discuss this in Section \ref{discussion}.

\subsection{Synthetic hateful posts through hateful language generation}
Next, we again augmented the existing 100 hateful posts using hateful language generated using the BLOOM large language model~\cite{scao2022bloom}, BLOOM-LM. In this method, we converted the entire hate speech dataset in the low-resource language into subsets of five posts and synthetically generate a sixth hateful post for each subset. We use the 100 available hateful posts in the low-resource language to generate 20 synthetic hateful posts. Then, we randomize the 100 posts in the low-resource language to re-order and re-group the hateful posts to form a new  prompt. This re-ordered dataset generates 20 more synthetic hateful posts. We iterate this step until we acquire the required number of synthetic posts. 

\emph{Finding.} We find that the BLOOM-LM method outperforms the MT method in both Hindi and Vietnamese as we increase the amount of synthetic data. BLOOM-LM also closes the gap in performance between the model trained on all original hateful posts of similar quantity as the BLOOM-augmented model. However, the CES method outperforms the BLOOM-LM method in Hindi in most cases while the BLOOM-LM method outperforms the CES method in most Vietnamese cases. Specifically, for Vietnamese, we observe that adding more BLOOM-LM synthetic data leads to a steady increase in performance. We hypothesize that this is potentially due to more representation of Vietnamese data in the BLOOM pretraining dataset compared to Hindi and discuss this in Section \ref{discussion}.

\subsection{Results on OOD test set}
To test the robustness of the CES method in comparison to the All-Orig, MT, and BLOOM-LM cases, we mimic a real-world deployment scenario and test the trained models on entirely new data from a different source than the training data. This is particularly challenging for hate speech models since differences in platform sources, hate lingo, narratives, etc can lead to entirely new forms of hate speech. We only found a different dataset for our OOD test in Hindi and thus use that for our analysis. 

\emph{Finding.} In the base case, training with 450 non-hateful and 100 original hateful, the mean F1 was 50.81. We observe that the BLOOM-LM method performs better than CES and MT methods on OOD data. As we incrementally add synthetic data, we noticed a reduction in performance for both MT and CES methods. MT on OOD test data dropped from a mean F1 of 45.85 to 41.49, and CES from 46.20 to 41.71 but for BLOOM-LM the performance ranged from 50.91 to 51.95. 

In general, we observe that in the OOD test, fewer training data performed better than more training data for all the methods---All-Orig, MT, CES, and BLOOM-LM. This makes sense since more data will increase the existing significant deviation between the training set and the new test set. Nonetheless, a CES approach may be more relevant for languages not represented in large language models like BLOOM. Since OOD data often represent the present state of the world at test/deployment time, we argue that incorporating newer entities from the real-world dataset into the entity table can significantly improve the performance of the CES method.

%% file: Sections/discussion.tex
\subsection{Interpretability analysis}
The performance boost obtained through training on synthetic data with the CES method helps validate our hypothesis of transferring hate speech context across languages. To develop a deeper understanding of our results and examine if the performance boost was, in fact, due to the presence of context-specific entities, we interpreted our model results using the SHapley Additive exPlanations (SHAP) framework \cite{lundberg2017unified}. The SHAP framework helps us calculate the contribution of each word when the model makes its prediction.

In our interpretability analysis, we obtained the average SHAP value for every word in the test data and sorted the words with maximum contribution across the entire dataset. We then annotated the top 20 words with respect to them being an entity or not and calculated the percentage contribution by entities across the top 20 contributing words in the test set. This analysis helps us understand whether the entities play a greater role in classifier prediction for the model trained on the synthetic data with contextual entity substitution (CES) when compared to that trained on machine-translated synthetic data (MT).

We observe that the average contribution of entities on classifier prediction is 31\% for the MT model. On the other hand, it is 38\% for the CES model in the Hindi language. We found even more promising results for Vietnamese as there was only a 13\% contribution by the entities towards the final prediction with MT while there was a 59\% contribution by entities in the CES model. This provides further evidence of entities playing a greater role in guiding the model prediction when the model is fine-tuned on the synthetic data with contextual entity substitution.

\subsection{Implications}
The scarcity of data for hate speech detection in low-resource language contexts has been well documented~\cite{fortuna2018survey, madukwe2020data, jahan2023systematic}. Data work for machine learning (hate speech detection inclusive) is considered boring, expensive, and intensive especially when accounting for geographic and language barriers~\cite{sambasivan2021everyone, nkemelu2022tackling}. Our work presents three significant implications: i) by presenting methods for augmenting hate speech data in limited data contexts and comparing their performance on in-domain and out-of-domain test sets, we address a lingering question for hate speech practitioners about technical approaches for boosting limited hate speech data for real-world deployments; 2) our empirical findings highlight the important role of humans-in-the-loop of hate speech detection systems for creating and maintaining structures, managing domain drifts, and evaluating performance and 3) we motivate the need for more research in synthetic hate speech data generation, and broadly, in the inclusion of more lower-resourced languages in large language models for use in downstream applications.

Our findings show that automatically translating hate speech data from one language is not the best approach for data augmentation. This is mostly due to the loss of contextual relevance of hate targets as the model translates from one language to another. Drawing from findings in vision systems~\cite{baradad2021learning}, two key properties that make synthetic data good is naturalism and diversity. Naturalism implies that the data may not be real but it must capture certain structural properties seen in real data. We attempt to achieve this natural property by translating data from one language to another. However, prior work has shown that machine translating hate speech data is subject to the quality of the translation system, the annotation scheme used in both languages, and class balance~\cite{casula2020hate}. 

Our contextual entity substitution method addresses a major limitation of machine-translated hate speech data by infusing structure and context into the translated results. The CES method also proffers the diversity property to the synthetic data generated. We have shown that this method outperforms simple machine translation and performs comparably to models trained using only original data or generative methods. However, since this method is heavily reliant on a finite set of entities in the entity table, we see no remarkable improvements as more CES synthetic data are generated. For instance, our analysis of the entity table in Hindi and Vietnamese from Table \ref{tab:ent_table} shows that the Hindi entity table having more target individuals than Vietnamese led to a more diverse synthetic data generation. The success of this method is dependent on the continuous update of the entity table to account for domain drifts and to improve the diversity of the generated synthetic data. This offers support to previous claims to include context experts as part of effective hate speech detection and tracking projects~\cite{nkemelu2022tackling}.

The BLOOM large language model used in this work has been trained on 46 natural languages~\cite{scao2022bloom} and our findings show that the level of language representation can play a role in the quality of the sentences generated by the model. For instance, since Vietnamese had twice the size of Hindi language data in the BLOOM pretraining dataset, we observe that the quality of synthetic data generated in Vietnamese is better than for Hindi. As models include more diverse languages in their pre-training setup, these methods can be extended to newer contexts. Future work can also explore the potential benefits of further finetuning the language models on data from the languages of interest prior to generating synthetic examples. Furthermore, our findings motivate the need for additional work in prompt engineering for synthetic hate speech data. Our initial experimentation with target-guided prompting (see Table \ref{tab:bloom_prompt_eng}) seeks to generate hate speech targeting a specific group. This approach could potentially improve the quality of synthetic data generated to train machine learning models.


Overall, we find that there is no single recipe for augmenting hate speech data in low-resource contexts. When the entity table is comprehensive, the CES method shines; when the language is well-represented in a generative large language model, the language generation technique performs well. In general, adapting hate speech from one context to another is bound to introduce noise and domain shift, choosing whether to perform contextual substitution or language generation will depend on the constraints in the limited data context of preference.

\subsection{Limitation and Future Work}
We recognize that this work presents some limitations, and some of them suggest promising directions for future work. Our present analyses have investigated the performance of our proposed methods on Hindi and Vietnamese even though these languages have reasonably decent representation in many language models. This selection bias might have influenced the performance of the proposed methods. Though within the context of hate speech detection research, there are very few resources in Hindi and Vietnamese~\cite{jahan2023systematic, joshi2020state}, it is unclear whether our methods will work for many less-resourced languages. We believe that approaches for hate speech detection using synthetic data generation should be extended to lesser-resourced languages and future work should consider this.

Our methodology adopts a random matching mechanism for selecting substituted entities from the table. Further work is needed to explore other matching methods. For example, exploring the effectiveness of adding another layer of semantic understanding to adapt the entities more closely to their corresponding replacement. This semantic coherence could potentially increase the quality of the hateful posts and further boost the performance of the models.

\begin{table}[b]
  \centering
  \caption{Target group-specific prompt engineering to generate targeted hateful posts using the BLOOM Language Model}
  \label{tab:bloom_prompt_eng}
  \begin{tabular}{|p{\linewidth}|}
  \hline
    \begin{center} \textbf{Prompt} \end{center}
    \textbf{Target group}: Muslim \textbf{Post}: If you are a Muslim, you are a criminal, end of the matter!\\
    \textbf{Target group}: Muslim \textbf{Post}: There is only one similarity between Kashmir and Palestine, Muslims are being openly oppressed in both the places and the world is silent!\\
    \textbf{Target group}: Muslim \textbf{Post}: The video of slogan of Pakistan Zindabad in the presence of Samajwadi Party MLA Abu Asim Azmi in Shramik Special train running from Wadala in Mumbai is going viral\\
    \textbf{Target group}: Muslim \textbf{Post}: If you are a Muslim then you are a criminal, and if you are anti-BJP then you are a straight terrorist.\\
    \textbf{Target group}: Muslim \textbf{Post}: In the case of rape and subsequent brutal murder of Dr. Priyanka Reddy in Hyderabad, India's so-called secularists are refraining from raising their voice today because the accused Muslim and the locality, Asaduddin Owaisi, is it not enough to protest?\\
    \textbf{Target group}: Muslim \textbf{Post}: \\
    \hline
    \begin{center}\textbf{Generated Post}\end{center}
    \textcolor{black}{If you are a Muslim or not a Hindu, you will have to leave the country.}\\
    \hline
  \end{tabular}
\end{table}


An ethical concern with researching hate speech detection methodologies using synthetically generated data is the possibility for bad actors to adopt these strategies for propagating synthetically generated hateful content on social media. While we unequivocally denounce such use, we argue that responsible use of the proposed methodologies can be deployed to inhibit the spread of such content from such malicious use. A model trained on synthetic data could even be more astute in detecting synthetic hate speech because of the distribution similarity with the data. The proposed methods could also be extended to incorporate techniques such as watermarking \cite{kirchenbauer2023watermark} to detect synthetically generated texts while retaining the benefits of data augmentation.



%% file: Sections/conclusion.tex
In this work, we address the issue of data imbalance and data unavailability affecting the performance of automatic hate speech detection systems in limited data contexts. We investigated three approaches to generate synthetic hate speech data and presented a novel methodology for transferring hateful sentiment across languages while retaining contextual relevance in the target domains.  We augmented a small number of hateful posts in Hindi and Vietnamese with synthetically generated hateful posts and trained machine learning models in a few-shot setup. Our findings show significant benefits of our proposed methods under different scenarios. Our contribution will help practitioners and researchers working on hate speech detection in limited data contexts build more robust machine learning systems to further their capacity to counter hate speech.



%% file: main.bbl

\begin{thebibliography}{71}


\ifx \showCODEN    \undefined \def \showCODEN     #1{\unskip}     \fi
\ifx \showDOI      \undefined \def \showDOI       #1{#1}\fi
\ifx \showISBNx    \undefined \def \showISBNx     #1{\unskip}     \fi
\ifx \showISBNxiii \undefined \def \showISBNxiii  #1{\unskip}     \fi
\ifx \showISSN     \undefined \def \showISSN      #1{\unskip}     \fi
\ifx \showLCCN     \undefined \def \showLCCN      #1{\unskip}     \fi
\ifx \shownote     \undefined \def \shownote      #1{#1}          \fi
\ifx \showarticletitle \undefined \def \showarticletitle #1{#1}   \fi
\ifx \showURL      \undefined \def \showURL       {\relax}        \fi
\providecommand\bibfield[2]{#2}
\providecommand\bibinfo[2]{#2}
\providecommand\natexlab[1]{#1}
\providecommand\showeprint[2][]{arXiv:#2}

\bibitem[pea(2021)]%
        {peacetechlab}
 \bibinfo{year}{2021}\natexlab{}.
\newblock \bibinfo{title}{{PeaceTech Lab | Hate Speech Lexicons}}.
\newblock
\newblock
\urldef\tempurl%
\url{https://www.peacetechlab.org/hate-speech}
\showURL{%
\tempurl}


\bibitem[Ali et~al\mbox{.}(2022)]%
        {ali2022hate}
\bibfield{author}{\bibinfo{person}{Raza Ali}, \bibinfo{person}{Umar Farooq}, \bibinfo{person}{Umair Arshad}, \bibinfo{person}{Waseem Shahzad}, {and} \bibinfo{person}{Mirza~Omer Beg}.} \bibinfo{year}{2022}\natexlab{}.
\newblock \showarticletitle{Hate speech detection on Twitter using transfer learning}.
\newblock \bibinfo{journal}{\emph{Computer Speech \& Language}}  \bibinfo{volume}{74} (\bibinfo{year}{2022}), \bibinfo{pages}{101365}.
\newblock


\bibitem[Alshalan et~al\mbox{.}(2020)]%
        {alshalan2020detection}
\bibfield{author}{\bibinfo{person}{Raghad Alshalan}, \bibinfo{person}{Hend Al-Khalifa}, \bibinfo{person}{Duaa Alsaeed}, \bibinfo{person}{Heyam Al-Baity}, {and} \bibinfo{person}{Shahad Alshalan}.} \bibinfo{year}{2020}\natexlab{}.
\newblock \showarticletitle{Detection of hate speech in covid-19--related tweets in the arab region: Deep learning and topic modeling approach}.
\newblock \bibinfo{journal}{\emph{Journal of Medical Internet Research}} \bibinfo{volume}{22}, \bibinfo{number}{12} (\bibinfo{year}{2020}), \bibinfo{pages}{e22609}.
\newblock


\bibitem[Aluru et~al\mbox{.}(2021)]%
        {aluru2021deep}
\bibfield{author}{\bibinfo{person}{Sai~Saketh Aluru}, \bibinfo{person}{Binny Mathew}, \bibinfo{person}{Punyajoy Saha}, {and} \bibinfo{person}{Animesh Mukherjee}.} \bibinfo{year}{2021}\natexlab{}.
\newblock \showarticletitle{A deep dive into multilingual hate speech classification}. In \bibinfo{booktitle}{\emph{Machine Learning and Knowledge Discovery in Databases. Applied Data Science and Demo Track: European Conference, ECML PKDD 2020, Ghent, Belgium, September 14--18, 2020, Proceedings, Part V}}. Springer, \bibinfo{pages}{423--439}.
\newblock


\bibitem[Badri et~al\mbox{.}(2022)]%
        {badri2022combining}
\bibfield{author}{\bibinfo{person}{Nabil Badri}, \bibinfo{person}{Ferihane Kboubi}, {and} \bibinfo{person}{Anja~Habacha Chaibi}.} \bibinfo{year}{2022}\natexlab{}.
\newblock \showarticletitle{Combining FastText and Glove word embedding for offensive and hate speech text detection}.
\newblock \bibinfo{journal}{\emph{Procedia Computer Science}}  \bibinfo{volume}{207} (\bibinfo{year}{2022}), \bibinfo{pages}{769--778}.
\newblock


\bibitem[Baradad~Jurjo et~al\mbox{.}(2021)]%
        {baradad2021learning}
\bibfield{author}{\bibinfo{person}{Manel Baradad~Jurjo}, \bibinfo{person}{Jonas Wulff}, \bibinfo{person}{Tongzhou Wang}, \bibinfo{person}{Phillip Isola}, {and} \bibinfo{person}{Antonio Torralba}.} \bibinfo{year}{2021}\natexlab{}.
\newblock \showarticletitle{Learning to see by looking at noise}.
\newblock \bibinfo{journal}{\emph{Advances in Neural Information Processing Systems}}  \bibinfo{volume}{34} (\bibinfo{year}{2021}), \bibinfo{pages}{2556--2569}.
\newblock


\bibitem[Bhardwaj et~al\mbox{.}(2020)]%
        {bhardwaj2020hostility}
\bibfield{author}{\bibinfo{person}{Mohit Bhardwaj}, \bibinfo{person}{Md~Shad Akhtar}, \bibinfo{person}{Asif Ekbal}, \bibinfo{person}{Amitava Das}, {and} \bibinfo{person}{Tanmoy Chakraborty}.} \bibinfo{year}{2020}\natexlab{}.
\newblock \showarticletitle{Hostility detection dataset in Hindi}.
\newblock \bibinfo{journal}{\emph{arXiv preprint arXiv:2011.03588}} (\bibinfo{year}{2020}).
\newblock


\bibitem[Bigoulaeva et~al\mbox{.}(2021)]%
        {bigoulaeva2021cross}
\bibfield{author}{\bibinfo{person}{Irina Bigoulaeva}, \bibinfo{person}{Viktor Hangya}, {and} \bibinfo{person}{Alexander Fraser}.} \bibinfo{year}{2021}\natexlab{}.
\newblock \showarticletitle{Cross-lingual transfer learning for hate speech detection}. In \bibinfo{booktitle}{\emph{Proceedings of the First Workshop on Language Technology for Equality, Diversity and Inclusion}}. \bibinfo{pages}{15--25}.
\newblock


\bibitem[Bohra et~al\mbox{.}(2018)]%
        {bohra2018dataset}
\bibfield{author}{\bibinfo{person}{Aditya Bohra}, \bibinfo{person}{Deepanshu Vijay}, \bibinfo{person}{Vinay Singh}, \bibinfo{person}{Syed~Sarfaraz Akhtar}, {and} \bibinfo{person}{Manish Shrivastava}.} \bibinfo{year}{2018}\natexlab{}.
\newblock \showarticletitle{A dataset of Hindi-English code-mixed social media text for hate speech detection}. In \bibinfo{booktitle}{\emph{Proceedings of the second workshop on computational modeling of people’s opinions, personality, and emotions in social media}}. \bibinfo{pages}{36--41}.
\newblock


\bibitem[Brown et~al\mbox{.}(2020)]%
        {brown2020language}
\bibfield{author}{\bibinfo{person}{Tom Brown}, \bibinfo{person}{Benjamin Mann}, \bibinfo{person}{Nick Ryder}, \bibinfo{person}{Melanie Subbiah}, \bibinfo{person}{Jared~D Kaplan}, \bibinfo{person}{Prafulla Dhariwal}, \bibinfo{person}{Arvind Neelakantan}, \bibinfo{person}{Pranav Shyam}, \bibinfo{person}{Girish Sastry}, \bibinfo{person}{Amanda Askell}, {et~al\mbox{.}}} \bibinfo{year}{2020}\natexlab{}.
\newblock \showarticletitle{Language models are few-shot learners}.
\newblock \bibinfo{journal}{\emph{Advances in neural information processing systems}}  \bibinfo{volume}{33} (\bibinfo{year}{2020}), \bibinfo{pages}{1877--1901}.
\newblock


\bibitem[Cao and Lee(2020)]%
        {cao2020hategan}
\bibfield{author}{\bibinfo{person}{Rui Cao} {and} \bibinfo{person}{Roy Ka-Wei Lee}.} \bibinfo{year}{2020}\natexlab{}.
\newblock \showarticletitle{HateGAN: Adversarial generative-based data augmentation for hate speech detection}. In \bibinfo{booktitle}{\emph{Proceedings of the 28th International Conference on Computational Linguistics}}. \bibinfo{pages}{6327--6338}.
\newblock


\bibitem[Cardoso et~al\mbox{.}(2022)]%
        {cardoso2022conditional}
\bibfield{author}{\bibinfo{person}{Renato Cardoso}, \bibinfo{person}{Sofia Vallecorsa}, {and} \bibinfo{person}{Edoardo Nemni}.} \bibinfo{year}{2022}\natexlab{}.
\newblock \showarticletitle{Conditional Progressive Generative Adversarial Network for satellite image generation}.
\newblock \bibinfo{journal}{\emph{arXiv preprint arXiv:2211.15303}} (\bibinfo{year}{2022}).
\newblock


\bibitem[Casula and Tonelli(2020)]%
        {casula2020hate}
\bibfield{author}{\bibinfo{person}{Camilla Casula} {and} \bibinfo{person}{Sara Tonelli}.} \bibinfo{year}{2020}\natexlab{}.
\newblock \showarticletitle{Hate speech detection with machine-translated data: the role of annotation scheme, class imbalance and undersampling}. In \bibinfo{booktitle}{\emph{Proceedings of the Seventh Italian Conference on Computational Linguistics, CLiC-it 2020}}, Vol.~\bibinfo{volume}{2769}. CEUR-WS. org.
\newblock


\bibitem[Chopra et~al\mbox{.}(2020)]%
        {chopra2020hindi}
\bibfield{author}{\bibinfo{person}{Shivang Chopra}, \bibinfo{person}{Ramit Sawhney}, \bibinfo{person}{Puneet Mathur}, {and} \bibinfo{person}{Rajiv~Ratn Shah}.} \bibinfo{year}{2020}\natexlab{}.
\newblock \showarticletitle{Hindi-english hate speech detection: Author profiling, debiasing, and practical perspectives}. In \bibinfo{booktitle}{\emph{Proceedings of the AAAI conference on artificial intelligence}}, Vol.~\bibinfo{volume}{34}. \bibinfo{pages}{386--393}.
\newblock


\bibitem[Conneau et~al\mbox{.}(2019)]%
        {conneau2019unsupervised}
\bibfield{author}{\bibinfo{person}{Alexis Conneau}, \bibinfo{person}{Kartikay Khandelwal}, \bibinfo{person}{Naman Goyal}, \bibinfo{person}{Vishrav Chaudhary}, \bibinfo{person}{Guillaume Wenzek}, \bibinfo{person}{Francisco Guzm{\'a}n}, \bibinfo{person}{Edouard Grave}, \bibinfo{person}{Myle Ott}, \bibinfo{person}{Luke Zettlemoyer}, {and} \bibinfo{person}{Veselin Stoyanov}.} \bibinfo{year}{2019}\natexlab{}.
\newblock \showarticletitle{Unsupervised cross-lingual representation learning at scale}.
\newblock \bibinfo{journal}{\emph{arXiv preprint arXiv:1911.02116}} (\bibinfo{year}{2019}).
\newblock


\bibitem[Das et~al\mbox{.}(2022)]%
        {das2022data}
\bibfield{author}{\bibinfo{person}{Mithun Das}, \bibinfo{person}{Somnath Banerjee}, {and} \bibinfo{person}{Animesh Mukherjee}.} \bibinfo{year}{2022}\natexlab{}.
\newblock \showarticletitle{Data bootstrapping approaches to improve low resource abusive language detection for indic languages}. In \bibinfo{booktitle}{\emph{Proceedings of the 33rd ACM Conference on Hypertext and Social Media}}. \bibinfo{pages}{32--42}.
\newblock


\bibitem[Davidson et~al\mbox{.}(2017)]%
        {davidson2017automated}
\bibfield{author}{\bibinfo{person}{Thomas Davidson}, \bibinfo{person}{Dana Warmsley}, \bibinfo{person}{Michael Macy}, {and} \bibinfo{person}{Ingmar Weber}.} \bibinfo{year}{2017}\natexlab{}.
\newblock \showarticletitle{Automated hate speech detection and the problem of offensive language}. In \bibinfo{booktitle}{\emph{Proceedings of the International AAAI Conference on Web and Social Media}}, Vol.~\bibinfo{volume}{11}. \bibinfo{pages}{512--515}.
\newblock


\bibitem[De~Gibert et~al\mbox{.}(2018)]%
        {de2018hate}
\bibfield{author}{\bibinfo{person}{Ona De~Gibert}, \bibinfo{person}{Naiara Perez}, \bibinfo{person}{Aitor Garc{\'\i}a-Pablos}, {and} \bibinfo{person}{Montse Cuadros}.} \bibinfo{year}{2018}\natexlab{}.
\newblock \showarticletitle{Hate speech dataset from a white supremacy forum}.
\newblock \bibinfo{journal}{\emph{arXiv preprint arXiv:1809.04444}} (\bibinfo{year}{2018}).
\newblock


\bibitem[Demilie and Salau(2022)]%
        {demilie2022detection}
\bibfield{author}{\bibinfo{person}{Wubetu~Barud Demilie} {and} \bibinfo{person}{Ayodeji~Olalekan Salau}.} \bibinfo{year}{2022}\natexlab{}.
\newblock \showarticletitle{Detection of fake news and hate speech for Ethiopian languages: a systematic review of the approaches}.
\newblock \bibinfo{journal}{\emph{Journal of big Data}} \bibinfo{volume}{9}, \bibinfo{number}{1} (\bibinfo{year}{2022}), \bibinfo{pages}{66}.
\newblock


\bibitem[Devlin et~al\mbox{.}(2018)]%
        {devlin2018bert}
\bibfield{author}{\bibinfo{person}{Jacob Devlin}, \bibinfo{person}{Ming-Wei Chang}, \bibinfo{person}{Kenton Lee}, {and} \bibinfo{person}{Kristina Toutanova}.} \bibinfo{year}{2018}\natexlab{}.
\newblock \showarticletitle{Bert: Pre-training of deep bidirectional transformers for language understanding}.
\newblock \bibinfo{journal}{\emph{arXiv preprint arXiv:1810.04805}} (\bibinfo{year}{2018}).
\newblock


\bibitem[Ding et~al\mbox{.}(2020)]%
        {Ding2020DAGADA}
\bibfield{author}{\bibinfo{person}{Bosheng Ding}, \bibinfo{person}{Linlin Liu}, \bibinfo{person}{Lidong Bing}, \bibinfo{person}{Canasai Kruengkrai}, \bibinfo{person}{Thien~Hai Nguyen}, \bibinfo{person}{Shafiq~R. Joty}, \bibinfo{person}{Luo Si}, {and} \bibinfo{person}{Chunyan Miao}.} \bibinfo{year}{2020}\natexlab{}.
\newblock \showarticletitle{DAGA: Data Augmentation with a Generation Approach for Low-resource Tagging Tasks}. In \bibinfo{booktitle}{\emph{Conference on Empirical Methods in Natural Language Processing}}.
\newblock


\bibitem[Fazel et~al\mbox{.}(2021)]%
        {fazel2021synthasr}
\bibfield{author}{\bibinfo{person}{Amin Fazel}, \bibinfo{person}{Wei Yang}, \bibinfo{person}{Yulan Liu}, \bibinfo{person}{Roberto Barra-Chicote}, \bibinfo{person}{Yixiong Meng}, \bibinfo{person}{Roland Maas}, {and} \bibinfo{person}{Jasha Droppo}.} \bibinfo{year}{2021}\natexlab{}.
\newblock \showarticletitle{Synthasr: Unlocking synthetic data for speech recognition}.
\newblock \bibinfo{journal}{\emph{arXiv preprint arXiv:2106.07803}} (\bibinfo{year}{2021}).
\newblock


\bibitem[Feng et~al\mbox{.}(2021a)]%
        {feng2021survey}
\bibfield{author}{\bibinfo{person}{Steven~Y Feng}, \bibinfo{person}{Varun Gangal}, \bibinfo{person}{Jason Wei}, \bibinfo{person}{Sarath Chandar}, \bibinfo{person}{Soroush Vosoughi}, \bibinfo{person}{Teruko Mitamura}, {and} \bibinfo{person}{Eduard Hovy}.} \bibinfo{year}{2021}\natexlab{a}.
\newblock \showarticletitle{A survey of data augmentation approaches for nlp}.
\newblock \bibinfo{journal}{\emph{arXiv preprint arXiv:2105.03075}} (\bibinfo{year}{2021}).
\newblock


\bibitem[Feng et~al\mbox{.}(2021b)]%
        {Feng2021ASO}
\bibfield{author}{\bibinfo{person}{Steven~Y. Feng}, \bibinfo{person}{Varun Gangal}, \bibinfo{person}{Jason Wei}, \bibinfo{person}{Sarath Chandar}, \bibinfo{person}{Soroush Vosoughi}, \bibinfo{person}{Teruko Mitamura}, {and} \bibinfo{person}{Eduard~H. Hovy}.} \bibinfo{year}{2021}\natexlab{b}.
\newblock \showarticletitle{A Survey of Data Augmentation Approaches for NLP}.
\newblock \bibinfo{journal}{\emph{ArXiv}}  \bibinfo{volume}{abs/2105.03075} (\bibinfo{year}{2021}).
\newblock


\bibitem[Fortuna and Nunes(2018)]%
        {fortuna2018survey}
\bibfield{author}{\bibinfo{person}{Paula Fortuna} {and} \bibinfo{person}{S{\'e}rgio Nunes}.} \bibinfo{year}{2018}\natexlab{}.
\newblock \showarticletitle{A survey on automatic detection of hate speech in text}.
\newblock \bibinfo{journal}{\emph{ACM Computing Surveys (CSUR)}} \bibinfo{volume}{51}, \bibinfo{number}{4} (\bibinfo{year}{2018}), \bibinfo{pages}{1--30}.
\newblock


\bibitem[Goodfellow et~al\mbox{.}(2020)]%
        {goodfellow2020generative}
\bibfield{author}{\bibinfo{person}{Ian Goodfellow}, \bibinfo{person}{Jean Pouget-Abadie}, \bibinfo{person}{Mehdi Mirza}, \bibinfo{person}{Bing Xu}, \bibinfo{person}{David Warde-Farley}, \bibinfo{person}{Sherjil Ozair}, \bibinfo{person}{Aaron Courville}, {and} \bibinfo{person}{Yoshua Bengio}.} \bibinfo{year}{2020}\natexlab{}.
\newblock \showarticletitle{Generative adversarial networks}.
\newblock \bibinfo{journal}{\emph{Commun. ACM}} \bibinfo{volume}{63}, \bibinfo{number}{11} (\bibinfo{year}{2020}), \bibinfo{pages}{139--144}.
\newblock


\bibitem[Gr{\"o}ndahl et~al\mbox{.}(2018)]%
        {grondahl2018all}
\bibfield{author}{\bibinfo{person}{Tommi Gr{\"o}ndahl}, \bibinfo{person}{Luca Pajola}, \bibinfo{person}{Mika Juuti}, \bibinfo{person}{Mauro Conti}, {and} \bibinfo{person}{N Asokan}.} \bibinfo{year}{2018}\natexlab{}.
\newblock \showarticletitle{All you need is" love" evading hate speech detection}. In \bibinfo{booktitle}{\emph{Proceedings of the 11th ACM workshop on artificial intelligence and security}}. \bibinfo{pages}{2--12}.
\newblock


\bibitem[Hartvigsen et~al\mbox{.}(2022)]%
        {Hartvigsen2022ToxiGenAL}
\bibfield{author}{\bibinfo{person}{Thomas Hartvigsen}, \bibinfo{person}{Saadia Gabriel}, \bibinfo{person}{Hamid Palangi}, \bibinfo{person}{Maarten Sap}, \bibinfo{person}{Dipankar Ray}, {and} \bibinfo{person}{Ece Kamar}.} \bibinfo{year}{2022}\natexlab{}.
\newblock \showarticletitle{ToxiGen: A Large-Scale Machine-Generated Dataset for Adversarial and Implicit Hate Speech Detection}.
\newblock \bibinfo{journal}{\emph{ArXiv}}  \bibinfo{volume}{abs/2203.09509} (\bibinfo{year}{2022}).
\newblock


\bibitem[Hu et~al\mbox{.}(2022)]%
        {hu2022synt++}
\bibfield{author}{\bibinfo{person}{Ting-Yao Hu}, \bibinfo{person}{Mohammadreza Armandpour}, \bibinfo{person}{Ashish Shrivastava}, \bibinfo{person}{Jen-Hao~Rick Chang}, \bibinfo{person}{Hema Koppula}, {and} \bibinfo{person}{Oncel Tuzel}.} \bibinfo{year}{2022}\natexlab{}.
\newblock \showarticletitle{Synt++: Utilizing imperfect synthetic data to improve speech recognition}. In \bibinfo{booktitle}{\emph{ICASSP 2022-2022 IEEE International Conference on Acoustics, Speech and Signal Processing (ICASSP)}}. IEEE, \bibinfo{pages}{7682--7686}.
\newblock


\bibitem[Jahan and Oussalah(2023)]%
        {jahan2023systematic}
\bibfield{author}{\bibinfo{person}{Md~Saroar Jahan} {and} \bibinfo{person}{Mourad Oussalah}.} \bibinfo{year}{2023}\natexlab{}.
\newblock \showarticletitle{A systematic review of Hate Speech automatic detection using Natural Language Processing.}
\newblock \bibinfo{journal}{\emph{Neurocomputing}} (\bibinfo{year}{2023}), \bibinfo{pages}{126232}.
\newblock


\bibitem[Joshi et~al\mbox{.}(2020)]%
        {joshi2020state}
\bibfield{author}{\bibinfo{person}{Pratik Joshi}, \bibinfo{person}{Sebastin Santy}, \bibinfo{person}{Amar Budhiraja}, \bibinfo{person}{Kalika Bali}, {and} \bibinfo{person}{Monojit Choudhury}.} \bibinfo{year}{2020}\natexlab{}.
\newblock \showarticletitle{The state and fate of linguistic diversity and inclusion in the NLP world}.
\newblock \bibinfo{journal}{\emph{arXiv preprint arXiv:2004.09095}} (\bibinfo{year}{2020}).
\newblock


\bibitem[Khemchandani et~al\mbox{.}(2021)]%
        {Khemchandani2021ExploitingLR}
\bibfield{author}{\bibinfo{person}{Yash Khemchandani}, \bibinfo{person}{Sarvesh Mehtani}, \bibinfo{person}{Vaidehi Patil}, \bibinfo{person}{Abhijeet Awasthi}, \bibinfo{person}{Partha~Pratim Talukdar}, {and} \bibinfo{person}{Sunita Sarawagi}.} \bibinfo{year}{2021}\natexlab{}.
\newblock \showarticletitle{Exploiting Language Relatedness for Low Web-Resource Language Model Adaptation: An Indic Languages Study}. In \bibinfo{booktitle}{\emph{Annual Meeting of the Association for Computational Linguistics}}.
\newblock


\bibitem[Kim(2022)]%
        {kim2022transferable}
\bibfield{author}{\bibinfo{person}{Yo-whan Kim}.} \bibinfo{year}{2022}\natexlab{}.
\newblock \emph{\bibinfo{title}{How Transferable are Video Representations Based on Synthetic Data?}}
\newblock \bibinfo{thesistype}{Ph.\,D. Dissertation}. \bibinfo{school}{Massachusetts Institute of Technology}.
\newblock


\bibitem[Kirchenbauer et~al\mbox{.}(2023)]%
        {kirchenbauer2023watermark}
\bibfield{author}{\bibinfo{person}{John Kirchenbauer}, \bibinfo{person}{Jonas Geiping}, \bibinfo{person}{Yuxin Wen}, \bibinfo{person}{Jonathan Katz}, \bibinfo{person}{Ian Miers}, {and} \bibinfo{person}{Tom Goldstein}.} \bibinfo{year}{2023}\natexlab{}.
\newblock \showarticletitle{A Watermark for Large Language Models}.
\newblock \bibinfo{journal}{\emph{arXiv preprint arXiv:2301.10226}} (\bibinfo{year}{2023}).
\newblock


\bibitem[Lauscher et~al\mbox{.}(2020)]%
        {Lauscher2020FromZT}
\bibfield{author}{\bibinfo{person}{Anne Lauscher}, \bibinfo{person}{Vinit Ravishankar}, \bibinfo{person}{Ivan Vulic}, {and} \bibinfo{person}{Goran Glavas}.} \bibinfo{year}{2020}\natexlab{}.
\newblock \showarticletitle{From Zero to Hero: On the Limitations of Zero-Shot Cross-Lingual Transfer with Multilingual Transformers}.
\newblock \bibinfo{journal}{\emph{ArXiv}}  \bibinfo{volume}{abs/2005.00633} (\bibinfo{year}{2020}).
\newblock


\bibitem[Lee et~al\mbox{.}(2021)]%
        {lee2021neural}
\bibfield{author}{\bibinfo{person}{Kenton Lee}, \bibinfo{person}{Kelvin Guu}, \bibinfo{person}{Luheng He}, \bibinfo{person}{Tim Dozat}, {and} \bibinfo{person}{Hyung~Won Chung}.} \bibinfo{year}{2021}\natexlab{}.
\newblock \showarticletitle{Neural data augmentation via example extrapolation}.
\newblock \bibinfo{journal}{\emph{arXiv preprint arXiv:2102.01335}} (\bibinfo{year}{2021}).
\newblock


\bibitem[Lundberg and Lee(2017)]%
        {lundberg2017unified}
\bibfield{author}{\bibinfo{person}{Scott~M Lundberg} {and} \bibinfo{person}{Su-In Lee}.} \bibinfo{year}{2017}\natexlab{}.
\newblock \showarticletitle{A unified approach to interpreting model predictions}.
\newblock \bibinfo{journal}{\emph{Advances in neural information processing systems}}  \bibinfo{volume}{30} (\bibinfo{year}{2017}).
\newblock


\bibitem[Luu et~al\mbox{.}(2021)]%
        {luu2021large}
\bibfield{author}{\bibinfo{person}{Son~T Luu}, \bibinfo{person}{Kiet~Van Nguyen}, {and} \bibinfo{person}{Ngan Luu-Thuy Nguyen}.} \bibinfo{year}{2021}\natexlab{}.
\newblock \showarticletitle{A large-scale dataset for hate speech detection on vietnamese social media texts}. In \bibinfo{booktitle}{\emph{Advances and Trends in Artificial Intelligence. Artificial Intelligence Practices: 34th International Conference on Industrial, Engineering and Other Applications of Applied Intelligent Systems, IEA/AIE 2021, Kuala Lumpur, Malaysia, July 26--29, 2021, Proceedings, Part I 34}}. Springer, \bibinfo{pages}{415--426}.
\newblock


\bibitem[Madukwe et~al\mbox{.}(2020)]%
        {madukwe2020data}
\bibfield{author}{\bibinfo{person}{Kosisochukwu Madukwe}, \bibinfo{person}{Xiaoying Gao}, {and} \bibinfo{person}{Bing Xue}.} \bibinfo{year}{2020}\natexlab{}.
\newblock \showarticletitle{In data we trust: A critical analysis of hate speech detection datasets}. In \bibinfo{booktitle}{\emph{Proceedings of the Fourth Workshop on Online Abuse and Harms}}. \bibinfo{pages}{150--161}.
\newblock


\bibitem[Madukwe et~al\mbox{.}(2022)]%
        {madukwe2022token}
\bibfield{author}{\bibinfo{person}{Kosisochukwu~Judith Madukwe}, \bibinfo{person}{Xiaoying Gao}, {and} \bibinfo{person}{Bing Xue}.} \bibinfo{year}{2022}\natexlab{}.
\newblock \showarticletitle{Token replacement-based data augmentation methods for hate speech detection}.
\newblock \bibinfo{journal}{\emph{World Wide Web}} (\bibinfo{year}{2022}), \bibinfo{pages}{1--22}.
\newblock


\bibitem[Mathew et~al\mbox{.}(2019)]%
        {mathew2019spread}
\bibfield{author}{\bibinfo{person}{Binny Mathew}, \bibinfo{person}{Ritam Dutt}, \bibinfo{person}{Pawan Goyal}, {and} \bibinfo{person}{Animesh Mukherjee}.} \bibinfo{year}{2019}\natexlab{}.
\newblock \showarticletitle{Spread of hate speech in online social media}. In \bibinfo{booktitle}{\emph{Proceedings of the 10th ACM conference on web science}}. \bibinfo{pages}{173--182}.
\newblock


\bibitem[Mathew et~al\mbox{.}(2020)]%
        {mathew2020hate}
\bibfield{author}{\bibinfo{person}{Binny Mathew}, \bibinfo{person}{Anurag Illendula}, \bibinfo{person}{Punyajoy Saha}, \bibinfo{person}{Soumya Sarkar}, \bibinfo{person}{Pawan Goyal}, {and} \bibinfo{person}{Animesh Mukherjee}.} \bibinfo{year}{2020}\natexlab{}.
\newblock \showarticletitle{Hate begets hate: A temporal study of hate speech}.
\newblock \bibinfo{journal}{\emph{Proceedings of the ACM on Human-Computer Interaction}} \bibinfo{volume}{4}, \bibinfo{number}{CSCW2} (\bibinfo{year}{2020}), \bibinfo{pages}{1--24}.
\newblock


\bibitem[Mathew et~al\mbox{.}(2021)]%
        {mathew2021hatexplain}
\bibfield{author}{\bibinfo{person}{Binny Mathew}, \bibinfo{person}{Punyajoy Saha}, \bibinfo{person}{Seid~Muhie Yimam}, \bibinfo{person}{Chris Biemann}, \bibinfo{person}{Pawan Goyal}, {and} \bibinfo{person}{Animesh Mukherjee}.} \bibinfo{year}{2021}\natexlab{}.
\newblock \showarticletitle{Hatexplain: A benchmark dataset for explainable hate speech detection}. In \bibinfo{booktitle}{\emph{Proceedings of the AAAI Conference on Artificial Intelligence}}, Vol.~\bibinfo{volume}{35}. \bibinfo{pages}{14867--14875}.
\newblock


\bibitem[Meng et~al\mbox{.}(2021)]%
        {Meng2021MixSpeechDA}
\bibfield{author}{\bibinfo{person}{Linghui Meng}, \bibinfo{person}{Jin Xu}, \bibinfo{person}{Xu Tan}, \bibinfo{person}{Jindong Wang}, \bibinfo{person}{Tao Qin}, {and} \bibinfo{person}{Bo Xu}.} \bibinfo{year}{2021}\natexlab{}.
\newblock \showarticletitle{MixSpeech: Data Augmentation for Low-Resource Automatic Speech Recognition}.
\newblock \bibinfo{journal}{\emph{ICASSP 2021 - 2021 IEEE International Conference on Acoustics, Speech and Signal Processing (ICASSP)}} (\bibinfo{year}{2021}), \bibinfo{pages}{7008--7012}.
\newblock


\bibitem[NKEMELU et~al\mbox{.}(2022)]%
        {nkemelu2022tackling}
\bibfield{author}{\bibinfo{person}{DANIEL NKEMELU}, \bibinfo{person}{HARSHIL SHAH}, \bibinfo{person}{IRFAN ESSA}, {and} \bibinfo{person}{MICHAEL~L BEST}.} \bibinfo{year}{2022}\natexlab{}.
\newblock \showarticletitle{Tackling Hate Speech in Low-resource Languages with Context Experts}.
\newblock  (\bibinfo{year}{2022}).
\newblock


\bibitem[Oord et~al\mbox{.}(2016)]%
        {oord2016wavenet}
\bibfield{author}{\bibinfo{person}{Aaron van~den Oord}, \bibinfo{person}{Sander Dieleman}, \bibinfo{person}{Heiga Zen}, \bibinfo{person}{Karen Simonyan}, \bibinfo{person}{Oriol Vinyals}, \bibinfo{person}{Alex Graves}, \bibinfo{person}{Nal Kalchbrenner}, \bibinfo{person}{Andrew Senior}, {and} \bibinfo{person}{Koray Kavukcuoglu}.} \bibinfo{year}{2016}\natexlab{}.
\newblock \showarticletitle{Wavenet: A generative model for raw audio}.
\newblock \bibinfo{journal}{\emph{arXiv preprint arXiv:1609.03499}} (\bibinfo{year}{2016}).
\newblock


\bibitem[Ousidhoum et~al\mbox{.}(2019)]%
        {ousidhoum2019multilingual}
\bibfield{author}{\bibinfo{person}{Nedjma Ousidhoum}, \bibinfo{person}{Zizheng Lin}, \bibinfo{person}{Hongming Zhang}, \bibinfo{person}{Yangqiu Song}, {and} \bibinfo{person}{Dit-Yan Yeung}.} \bibinfo{year}{2019}\natexlab{}.
\newblock \showarticletitle{Multilingual and multi-aspect hate speech analysis}.
\newblock \bibinfo{journal}{\emph{arXiv preprint arXiv:1908.11049}} (\bibinfo{year}{2019}).
\newblock


\bibitem[Perez and Wang(2017)]%
        {Perez2017TheEO}
\bibfield{author}{\bibinfo{person}{Luis Perez} {and} \bibinfo{person}{Jason Wang}.} \bibinfo{year}{2017}\natexlab{}.
\newblock \showarticletitle{The Effectiveness of Data Augmentation in Image Classification using Deep Learning}.
\newblock \bibinfo{journal}{\emph{ArXiv}}  \bibinfo{volume}{abs/1712.04621} (\bibinfo{year}{2017}).
\newblock


\bibitem[Raffel et~al\mbox{.}(2020)]%
        {raffel2020exploring}
\bibfield{author}{\bibinfo{person}{Colin Raffel}, \bibinfo{person}{Noam Shazeer}, \bibinfo{person}{Adam Roberts}, \bibinfo{person}{Katherine Lee}, \bibinfo{person}{Sharan Narang}, \bibinfo{person}{Michael Matena}, \bibinfo{person}{Yanqi Zhou}, \bibinfo{person}{Wei Li}, {and} \bibinfo{person}{Peter~J Liu}.} \bibinfo{year}{2020}\natexlab{}.
\newblock \showarticletitle{Exploring the limits of transfer learning with a unified text-to-text transformer}.
\newblock \bibinfo{journal}{\emph{The Journal of Machine Learning Research}} \bibinfo{volume}{21}, \bibinfo{number}{1} (\bibinfo{year}{2020}), \bibinfo{pages}{5485--5551}.
\newblock


\bibitem[Ramesh et~al\mbox{.}(2022)]%
        {ramesh2022hierarchical}
\bibfield{author}{\bibinfo{person}{Aditya Ramesh}, \bibinfo{person}{Prafulla Dhariwal}, \bibinfo{person}{Alex Nichol}, \bibinfo{person}{Casey Chu}, {and} \bibinfo{person}{Mark Chen}.} \bibinfo{year}{2022}\natexlab{}.
\newblock \showarticletitle{Hierarchical text-conditional image generation with clip latents}.
\newblock \bibinfo{journal}{\emph{arXiv preprint arXiv:2204.06125}} (\bibinfo{year}{2022}).
\newblock


\bibitem[Ranasinghe and Zampieri(2021)]%
        {ranasinghe2021multilingual}
\bibfield{author}{\bibinfo{person}{Tharindu Ranasinghe} {and} \bibinfo{person}{Marcos Zampieri}.} \bibinfo{year}{2021}\natexlab{}.
\newblock \showarticletitle{Multilingual offensive language identification for low-resource languages}.
\newblock \bibinfo{journal}{\emph{Transactions on Asian and Low-Resource Language Information Processing}} \bibinfo{volume}{21}, \bibinfo{number}{1} (\bibinfo{year}{2021}), \bibinfo{pages}{1--13}.
\newblock


\bibitem[Research(2019)]%
        {github_fb_laser}
\bibfield{author}{\bibinfo{person}{Facebook Research}.} \bibinfo{year}{2019}\natexlab{}.
\newblock \bibinfo{booktitle}{\emph{LASER: Language-Agnostic SEntence Representations}}.
\newblock


\bibitem[Romim et~al\mbox{.}(2021)]%
        {romim2021hate}
\bibfield{author}{\bibinfo{person}{Nauros Romim}, \bibinfo{person}{Mosahed Ahmed}, \bibinfo{person}{Hriteshwar Talukder}, {and} \bibinfo{person}{Md Saiful~Islam}.} \bibinfo{year}{2021}\natexlab{}.
\newblock \showarticletitle{Hate speech detection in the bengali language: A dataset and its baseline evaluation}. In \bibinfo{booktitle}{\emph{Proceedings of International Joint Conference on Advances in Computational Intelligence: IJCACI 2020}}. Springer, \bibinfo{pages}{457--468}.
\newblock


\bibitem[Sagers et~al\mbox{.}(2022)]%
        {sagers2022improving}
\bibfield{author}{\bibinfo{person}{Luke~W Sagers}, \bibinfo{person}{James~A Diao}, \bibinfo{person}{Matthew Groh}, \bibinfo{person}{Pranav Rajpurkar}, \bibinfo{person}{Adewole~S Adamson}, {and} \bibinfo{person}{Arjun~K Manrai}.} \bibinfo{year}{2022}\natexlab{}.
\newblock \showarticletitle{Improving dermatology classifiers across populations using images generated by large diffusion models}.
\newblock \bibinfo{journal}{\emph{arXiv preprint arXiv:2211.13352}} (\bibinfo{year}{2022}).
\newblock


\bibitem[Saharia et~al\mbox{.}(2022)]%
        {saharia2022photorealistic}
\bibfield{author}{\bibinfo{person}{Chitwan Saharia}, \bibinfo{person}{William Chan}, \bibinfo{person}{Saurabh Saxena}, \bibinfo{person}{Lala Li}, \bibinfo{person}{Jay Whang}, \bibinfo{person}{Emily Denton}, \bibinfo{person}{Seyed Kamyar~Seyed Ghasemipour}, \bibinfo{person}{Burcu~Karagol Ayan}, \bibinfo{person}{S~Sara Mahdavi}, \bibinfo{person}{Rapha~Gontijo Lopes}, {et~al\mbox{.}}} \bibinfo{year}{2022}\natexlab{}.
\newblock \showarticletitle{Photorealistic text-to-image diffusion models with deep language understanding}.
\newblock \bibinfo{journal}{\emph{arXiv preprint arXiv:2205.11487}} (\bibinfo{year}{2022}).
\newblock


\bibitem[Sambasivan et~al\mbox{.}(2021)]%
        {sambasivan2021everyone}
\bibfield{author}{\bibinfo{person}{Nithya Sambasivan}, \bibinfo{person}{Shivani Kapania}, \bibinfo{person}{Hannah Highfill}, \bibinfo{person}{Diana Akrong}, \bibinfo{person}{Praveen Paritosh}, {and} \bibinfo{person}{Lora~M Aroyo}.} \bibinfo{year}{2021}\natexlab{}.
\newblock \showarticletitle{“Everyone wants to do the model work, not the data work”: Data Cascades in High-Stakes AI}. In \bibinfo{booktitle}{\emph{proceedings of the 2021 CHI Conference on Human Factors in Computing Systems}}. \bibinfo{pages}{1--15}.
\newblock


\bibitem[Scao et~al\mbox{.}(2022)]%
        {scao2022bloom}
\bibfield{author}{\bibinfo{person}{Teven~Le Scao}, \bibinfo{person}{Angela Fan}, \bibinfo{person}{Christopher Akiki}, \bibinfo{person}{Ellie Pavlick}, \bibinfo{person}{Suzana Ili{\'c}}, \bibinfo{person}{Daniel Hesslow}, \bibinfo{person}{Roman Castagn{\'e}}, \bibinfo{person}{Alexandra~Sasha Luccioni}, \bibinfo{person}{Fran{\c{c}}ois Yvon}, \bibinfo{person}{Matthias Gall{\'e}}, {et~al\mbox{.}}} \bibinfo{year}{2022}\natexlab{}.
\newblock \showarticletitle{Bloom: A 176b-parameter open-access multilingual language model}.
\newblock \bibinfo{journal}{\emph{arXiv preprint arXiv:2211.05100}} (\bibinfo{year}{2022}).
\newblock


\bibitem[Schmidt and Wiegand(2019)]%
        {schmidt2019survey}
\bibfield{author}{\bibinfo{person}{Anna Schmidt} {and} \bibinfo{person}{Michael Wiegand}.} \bibinfo{year}{2019}\natexlab{}.
\newblock \showarticletitle{A survey on hate speech detection using natural language processing}. In \bibinfo{booktitle}{\emph{Proceedings of the Fifth International Workshop on Natural Language Processing for Social Media, April 3, 2017, Valencia, Spain}}. Association for Computational Linguistics, \bibinfo{pages}{1--10}.
\newblock


\bibitem[Shen et~al\mbox{.}(2018)]%
        {shen2018natural}
\bibfield{author}{\bibinfo{person}{Jonathan Shen}, \bibinfo{person}{Ruoming Pang}, \bibinfo{person}{Ron~J Weiss}, \bibinfo{person}{Mike Schuster}, \bibinfo{person}{Navdeep Jaitly}, \bibinfo{person}{Zongheng Yang}, \bibinfo{person}{Zhifeng Chen}, \bibinfo{person}{Yu Zhang}, \bibinfo{person}{Yuxuan Wang}, \bibinfo{person}{Rj Skerrv-Ryan}, {et~al\mbox{.}}} \bibinfo{year}{2018}\natexlab{}.
\newblock \showarticletitle{Natural tts synthesis by conditioning wavenet on mel spectrogram predictions}. In \bibinfo{booktitle}{\emph{2018 IEEE international conference on acoustics, speech and signal processing (ICASSP)}}. IEEE, \bibinfo{pages}{4779--4783}.
\newblock


\bibitem[Shorten and Khoshgoftaar(2019)]%
        {shorten2019survey}
\bibfield{author}{\bibinfo{person}{Connor Shorten} {and} \bibinfo{person}{Taghi~M Khoshgoftaar}.} \bibinfo{year}{2019}\natexlab{}.
\newblock \showarticletitle{A survey on image data augmentation for deep learning}.
\newblock \bibinfo{journal}{\emph{Journal of big data}} \bibinfo{volume}{6}, \bibinfo{number}{1} (\bibinfo{year}{2019}), \bibinfo{pages}{1--48}.
\newblock


\bibitem[Swamy et~al\mbox{.}(2019)]%
        {swamy2019studying}
\bibfield{author}{\bibinfo{person}{Steve~Durairaj Swamy}, \bibinfo{person}{Anupam Jamatia}, {and} \bibinfo{person}{Bj{\"o}rn Gamb{\"a}ck}.} \bibinfo{year}{2019}\natexlab{}.
\newblock \showarticletitle{Studying generalisability across abusive language detection datasets}. In \bibinfo{booktitle}{\emph{Proceedings of the 23rd conference on computational natural language learning (CoNLL)}}. \bibinfo{pages}{940--950}.
\newblock


\bibitem[Thoppilan et~al\mbox{.}(2022)]%
        {thoppilan2022lamda}
\bibfield{author}{\bibinfo{person}{Romal Thoppilan}, \bibinfo{person}{Daniel De~Freitas}, \bibinfo{person}{Jamie Hall}, \bibinfo{person}{Noam Shazeer}, \bibinfo{person}{Apoorv Kulshreshtha}, \bibinfo{person}{Heng-Tze Cheng}, \bibinfo{person}{Alicia Jin}, \bibinfo{person}{Taylor Bos}, \bibinfo{person}{Leslie Baker}, \bibinfo{person}{Yu Du}, {et~al\mbox{.}}} \bibinfo{year}{2022}\natexlab{}.
\newblock \showarticletitle{Lamda: Language models for dialog applications}.
\newblock \bibinfo{journal}{\emph{arXiv preprint arXiv:2201.08239}} (\bibinfo{year}{2022}).
\newblock


\bibitem[Toraman et~al\mbox{.}(2022)]%
        {toraman2022large}
\bibfield{author}{\bibinfo{person}{Cagri Toraman}, \bibinfo{person}{Furkan {\c{S}}ahinu{\c{c}}}, {and} \bibinfo{person}{Eyup~Halit Y{\i}lmaz}.} \bibinfo{year}{2022}\natexlab{}.
\newblock \showarticletitle{Large-scale hate speech detection with cross-domain transfer}.
\newblock \bibinfo{journal}{\emph{arXiv preprint arXiv:2203.01111}} (\bibinfo{year}{2022}).
\newblock


\bibitem[Wei et~al\mbox{.}(2020)]%
        {wei2020comparison}
\bibfield{author}{\bibinfo{person}{Shengyun Wei}, \bibinfo{person}{Shun Zou}, \bibinfo{person}{Feifan Liao}, {et~al\mbox{.}}} \bibinfo{year}{2020}\natexlab{}.
\newblock \showarticletitle{A comparison on data augmentation methods based on deep learning for audio classification}. In \bibinfo{booktitle}{\emph{Journal of Physics: Conference Series}}, Vol.~\bibinfo{volume}{1453}. IOP Publishing, \bibinfo{pages}{012085}.
\newblock


\bibitem[Whang et~al\mbox{.}(2023)]%
        {whang2023data}
\bibfield{author}{\bibinfo{person}{Steven~Euijong Whang}, \bibinfo{person}{Yuji Roh}, \bibinfo{person}{Hwanjun Song}, {and} \bibinfo{person}{Jae-Gil Lee}.} \bibinfo{year}{2023}\natexlab{}.
\newblock \showarticletitle{Data collection and quality challenges in deep learning: A data-centric ai perspective}.
\newblock \bibinfo{journal}{\emph{The VLDB Journal}} (\bibinfo{year}{2023}), \bibinfo{pages}{1--23}.
\newblock


\bibitem[Wu and Dredze(2020)]%
        {wu2020all}
\bibfield{author}{\bibinfo{person}{Shijie Wu} {and} \bibinfo{person}{Mark Dredze}.} \bibinfo{year}{2020}\natexlab{}.
\newblock \showarticletitle{Are all languages created equal in multilingual BERT?}
\newblock \bibinfo{journal}{\emph{arXiv preprint arXiv:2005.09093}} (\bibinfo{year}{2020}).
\newblock


\bibitem[Xia et~al\mbox{.}(2019)]%
        {Xia2019GeneralizedDA}
\bibfield{author}{\bibinfo{person}{M. Xia}, \bibinfo{person}{X. Kong}, \bibinfo{person}{Antonios Anastasopoulos}, {and} \bibinfo{person}{Graham Neubig}.} \bibinfo{year}{2019}\natexlab{}.
\newblock \showarticletitle{Generalized Data Augmentation for Low-Resource Translation}.
\newblock \bibinfo{journal}{\emph{ArXiv}}  \bibinfo{volume}{abs/1906.03785} (\bibinfo{year}{2019}).
\newblock


\bibitem[Yoder et~al\mbox{.}(2022)]%
        {yoder2022hate}
\bibfield{author}{\bibinfo{person}{Michael~Miller Yoder}, \bibinfo{person}{Lynnette Hui~Xian Ng}, \bibinfo{person}{David~West Brown}, {and} \bibinfo{person}{Kathleen~M Carley}.} \bibinfo{year}{2022}\natexlab{}.
\newblock \showarticletitle{How Hate Speech Varies by Target Identity: A Computational Analysis}.
\newblock \bibinfo{journal}{\emph{arXiv preprint arXiv:2210.10839}} (\bibinfo{year}{2022}).
\newblock


\bibitem[Yu et~al\mbox{.}(2022)]%
        {yu2022scaling}
\bibfield{author}{\bibinfo{person}{Jiahui Yu}, \bibinfo{person}{Yuanzhong Xu}, \bibinfo{person}{Jing~Yu Koh}, \bibinfo{person}{Thang Luong}, \bibinfo{person}{Gunjan Baid}, \bibinfo{person}{Zirui Wang}, \bibinfo{person}{Vijay Vasudevan}, \bibinfo{person}{Alexander Ku}, \bibinfo{person}{Yinfei Yang}, \bibinfo{person}{Burcu~Karagol Ayan}, {et~al\mbox{.}}} \bibinfo{year}{2022}\natexlab{}.
\newblock \showarticletitle{Scaling autoregressive models for content-rich text-to-image generation}.
\newblock \bibinfo{journal}{\emph{arXiv preprint arXiv:2206.10789}} (\bibinfo{year}{2022}).
\newblock


\bibitem[Yun et~al\mbox{.}(2020)]%
        {yun2020videomix}
\bibfield{author}{\bibinfo{person}{Sangdoo Yun}, \bibinfo{person}{Seong~Joon Oh}, \bibinfo{person}{Byeongho Heo}, \bibinfo{person}{Dongyoon Han}, {and} \bibinfo{person}{Jinhyung Kim}.} \bibinfo{year}{2020}\natexlab{}.
\newblock \showarticletitle{Videomix: Rethinking data augmentation for video classification}.
\newblock \bibinfo{journal}{\emph{arXiv preprint arXiv:2012.03457}} (\bibinfo{year}{2020}).
\newblock


\bibitem[Ziqi et~al\mbox{.}(2019)]%
        {ziqi2019hate}
\bibfield{author}{\bibinfo{person}{Z Ziqi}, \bibinfo{person}{D Robinson}, {and} \bibinfo{person}{T Jonathan}.} \bibinfo{year}{2019}\natexlab{}.
\newblock \showarticletitle{Hate speech detection using a convolution-LSTM based deep neural network}.
\newblock \bibinfo{journal}{\emph{IJCCS}}  \bibinfo{volume}{11816} (\bibinfo{year}{2019}), \bibinfo{pages}{2546--2553}.
\newblock


\end{thebibliography}
